\documentclass[10pt,twocolumn,letterpaper]{article}

\usepackage{cvpr}              

\usepackage{graphicx}
\usepackage{amsmath}
\usepackage{amssymb}
\usepackage{booktabs}

\usepackage[table]{xcolor}

\usepackage[misc]{ifsym}

\usepackage[pagebackref,breaklinks,colorlinks]{hyperref}

\usepackage[capitalize]{cleveref}
\crefname{section}{Sec.}{Secs.}
\Crefname{section}{Section}{Sections}
\Crefname{table}{Table}{Tables}
\crefname{table}{Tab.}{Tabs.}

\def\cvprPaperID{139} 
\def\confName{CVPR}
\def\confYear{2022}

\begin{document}

\title{Weakly Supervised Semantic Segmentation by Pixel-to-Prototype Contrast}

\author{Ye Du,\ \  Zehua Fu,\ \  Qingjie Liu\footnotemark[1],\ \  Yunhong Wang\\
State Key Laboratory of Virtual Reality Technology and Systems, Beihang University, Beijing, China\\
Hangzhou Innovation Institute, Beihang University\\
{\tt\small \{duyee, zehua\_fu, qingjie.liu, yhwang\}@buaa.edu.cn}
}


\maketitle

\renewcommand{\thefootnote}{\fnsymbol{footnote}} 
\footnotetext[1]{Corresponding author.} 

\begin{abstract}
  Though image-level weakly supervised semantic segmentation (WSSS) has achieved great progress with Class Activation Maps (CAMs) as the cornerstone, the large supervision gap between classification and segmentation still hampers the model to generate more complete and precise pseudo masks for segmentation.
  In this study, we propose weakly-supervised pixel-to-prototype contrast that can provide pixel-level supervisory signals to narrow the gap.
  Guided by two intuitive priors, our method is executed across different views and within per single view of an image, aiming to impose cross-view feature semantic consistency regularization and facilitate intra(inter)-class compactness(dispersion) of the feature space.
  Our method can be seamlessly incorporated into existing WSSS models without any changes to the base networks and does not incur any extra inference burden. 
  Extensive experiments manifest that our method consistently improves two strong baselines by large margins, demonstrating the effectiveness.
  Specifically, built on top of SEAM, we improve the initial seed mIoU on PASCAL VOC 2012 from 55.4\% to 61.5\%.
  Moreover, armed with our method, we increase the segmentation mIoU of EPS from 70.8\% to 73.6\%, achieving new state-of-the-art. 
\end{abstract}

\section{Introduction}
\label{sec:intro}

Benefiting from large-scale pixel-level annotations, semantic segmentation \cite{minaee2021imagesegmentation} has achieved remarkable progress in recent years. However, obtaining such precise pixel-wise annotations is laborious and time-consuming. To relieve this burden, many works resort to weakly supervised semantic segmentation (WSSS) that aims at learning segmentation models from weak labels such as image tags \cite{kolesnikov2016SEC, ahn2018learningAffinityNet, jiang2019integralOAA, zhang2020causalCONTA, wang2020selfSEAM, SPML, lee2021railroadEPS}, bounding boxes \cite{oh2021backgroundBox2}, points \cite{bearman2016sPoints1} and scribbles \cite{tang2018regularizedScrible2}. 
Among them, image-level WSSS that requires only image tags has been extensively studied in the computer vision community.

\begin{figure}[t]
\centering
\begin{center}
\includegraphics[width=0.45\textwidth]{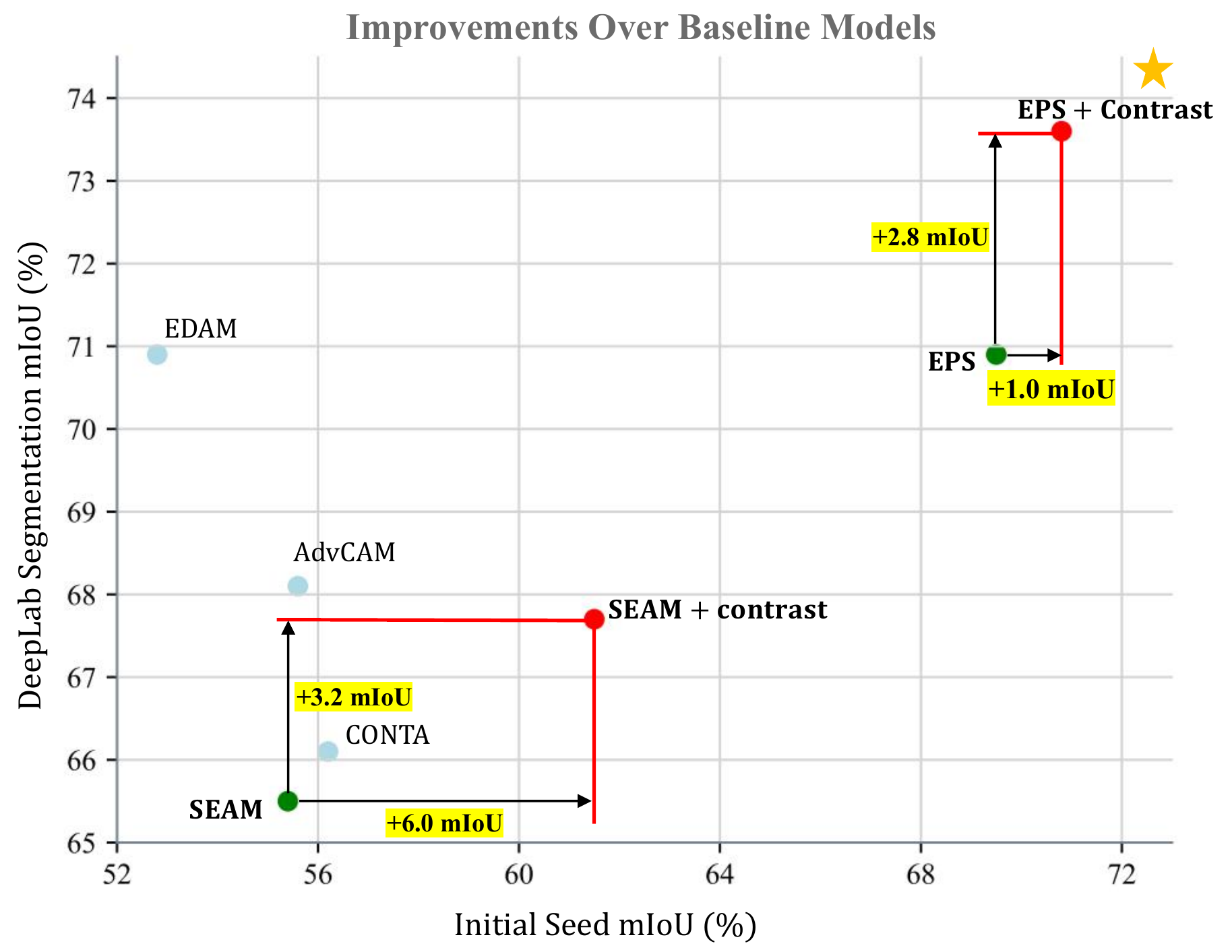}
\end{center}
\caption{Initial seed quality \textit{vs.} segmentation performance. Our method enables consistent performance improvements over state-of-the-arts, \textit{i.e.}, SEAM \cite{wang2020selfSEAM} and EPS \cite{lee2021railroadEPS}, without bringing any changes to the base networks during inference.}
\label{fig:plot_performance}
\end{figure}

Image-level WSSS is a challenging task since the image tags indicate only the existence of object categories and do not inform accurate object locations that are essential for semantic segmentation. 
To tackle this issue, Class Activation Maps (CAMs) \cite{zhou2016learningCAM} that identify which parts of the image contribute the most to the classification have been widely adopted to roughly estimate the regions of the target objects.
The regions, also known as seeds, are used to generate pseudo ground truths for training segmentation models.
However, CAMs only cover parts of objects, resulting in inaccurate and incomplete supervision.
This issue is rooted in the supervision gap between the classification and the segmentation tasks.
Specifically, classification networks supervised by image tags tend to focus on the most discriminative regions of objects for achieving a better performance of correctly tagging the image, while the segmentation task requires pixel-level supervision to assign a category to each pixel within the whole image.
Narrowing down the supervision gap is crucial for WSSS, motivating us to explore pixel-wise supervisory signals that are complementary to image tags. 

Inspired by the compelling contrastive self-supervised algorithms \cite{jaiswal2021CLsurvey1}, we develop a novel \textit{weakly-supervised pixel-to-prototype contrastive learning} method for WSSS, which can provide pixel-level supervision to improve the quality of CAMs and the performance of segmentation.
Our method is based on two implicit but valuable priors:
\textbf{(i)} features should retain semantic consistency across different views of an image; and
\textbf{(ii)} pixels sharing the same label should have similar representations in the feature space, and vice versa.
With these priors as guidelines, the pixel-to-prototype contrast is executed across different views and within per single view of each image, respectively, leading us to the cross-view contrast and intra-view contrast.

Our method is instantiated with a unified pixel-to-prototype contrastive learning formulation, which shapes the pixel embedding space through a prototype-based metric learning methodology. 
The core idea is pulling pixels together to their positive prototypes and pushing them away from their negative prototypes to learn discriminative dense visual representations. 
In our method, a prototype is defined as a representative embedding of a category. It is estimated from pixel-wise feature embeddings with the top
activations in the CAMs. 
During learning, the polarity of each prototype is determined by the pseudo label of each pixel coupled to it in the current mini-batch. 
However, when generating pseudo masks from CAMs, there emerges a tricky problem: the over-activated and under-activated regions may corrupt the contrastive learning, especially the intra-view contrast.
To alleviate this issue, we adopt two strategies: semi-hard prototype mining and hard pixel sampling to reduce the inaccurate contrasts as well as better utilize hard samples.

Recently, Wang \textit{et al.} \cite{wang2020selfSEAM} propose SEAM to mitigate the supervision gap issue with a CAM equivariance constraint that enforces the CAMs to have the same spatial transformation as the input images. Our method has two major differences.
Firstly, our method imposes regularizations to pixel-level features, enforcing the pixel embedding to be similar to the positive prototype and dissimilar to negative prototypes, while SEAM calculates consistency loss between CAMs of different views of the same image. 
Moreover, both cross-view and intra-view regularizations are considered in our work, while SEAM only integrates equivariant regularization across views.

Our method can be seamlessly incorporated into existing WSSS models without any changes to the base networks.
It requires only additional common projectors during training and does not incur an extra inference burden.
Experiments demonstrate that our method improves state-of-the-art models by large margins.
As shown in \cref{fig:plot_performance}, our method consistently improves two strong baseline models \textit{w.r.t.} both initial seed qualities and segmentation performance.
We also validate our method through an extensive ablation study, where we find that each component contributes substantial performance improvements.

To summary, our main contributions are as follows: 
\begin{itemize}
\item We propose weakly-supervised pixel-to-prototype contrast for WSSS. It enables the pixels to receive supervision from the reliable prototype of each class under the WSSS setting, which substantially narrows the gap between classification and segmentation. 
\item We propose to perform pixel-to-prototype contrastive learning both within a view and across different views of an image, which significantly improves the qualities of CAMs and the subsequent segmentation masks. 
\item Our method shows impressive results, surpassing baseline models by large margins and achieving the top performance on the standard benchmark.
\end{itemize}

 \section{Related Work}
\noindent \textbf{Image-level WSSS.}
WSSS with image-level labels has achieved significant progress under the pipeline of first generating pseudo masks then training a semantic segmentation network.
Recent methods for WSSS rely on CAMs \cite{zhou2016learningCAM} to inform object locations by discovering image pixels that are informative for classification.
However, CAMs can only highlight the most discriminative regions of objects, thus offering incomplete pseudo masks.
Substantial efforts have been devoted to resolving this issue.
They intend to complete the CAMs by enforcing networks to pay more attention to non-discriminative object regions using strategies such as region erasing \cite{hou2018selfErasing, wei2017objectERASING}, region supervision \cite{kim2021discriminativeDRS}, and region growing \cite{huang2018weaklyRegionGrowing, shimoda2019self-supervised-difference}.
Some other works refine the CAM by an iterative solution.
For instance, PSA \cite{ahn2018learningAffinityNet} and IRN \cite{ahn2019weaklyIRNet} propose to propagate local responses to nearby areas which belong to the same semantic entity via random walk.

The inherent reason for the above issue is due to the supervision gap between classification and segmentation.
Noticing this, many researchers explore to use additional supervision, such as multi-level feature maps \cite{kim2016deconvolutionalmulti-feature}, accumulated feature maps \cite{jiang2019integralOAA}, cross-image semantics \cite{fan2020cianCIAN, sun2020miningcrossimage}, sub-categories \cite{chang2020weaklySubCategory}, saliency maps \cite{yao2020saliency2, lee2021railroadEPS}, and CAM consistency constraints \cite{wang2020selfSEAM}, to narrow the gap.
These approaches are simple but achieve encouraging performance.


\begin{figure*}[htp]
\centering
\begin{center}
\includegraphics[width=0.9\textwidth]{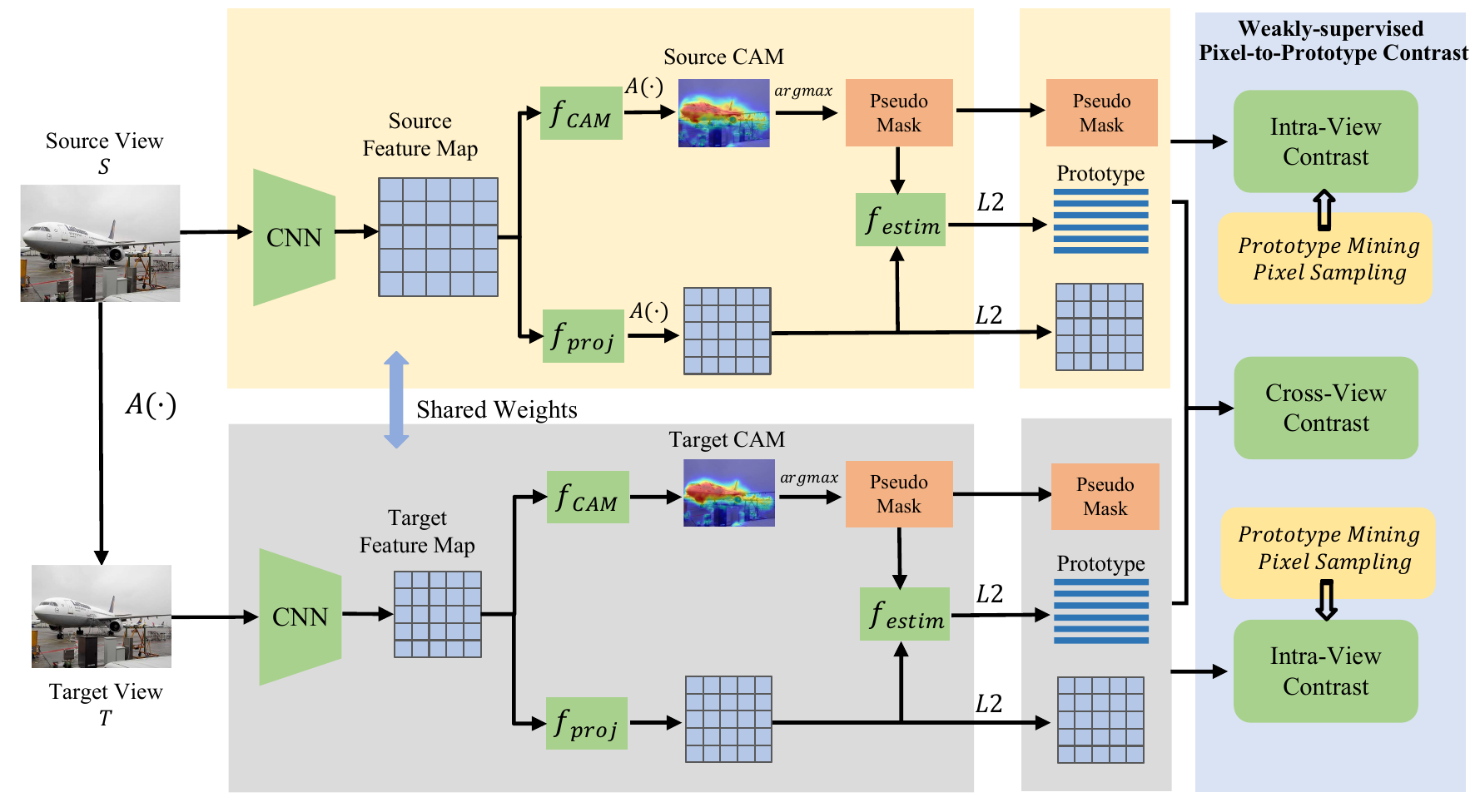}
\end{center}
\caption{The overall pipeline of our proposed pixel-to-prototype contrast for WSSS. $A(\cdot)$ is a spatial transformation for augmenting training samples. $f_{CAM}$, $f_{proj}$ are implemented by 1×1 convolutional layer followed by ReLU. $f_{est}$ represents the prototype estimation process and $\boldsymbol{p}^{\{S,T\}}$ represent the generated prototypes. \textit{L2} denotes per-pixel L2 normalization. The \textit{argmax} function is conducted per-pixel along the channel dimension and returns the index of the maximum value.}
\label{overall}
\end{figure*}

\noindent \textbf{Contrastive Learning.}
Contrastive learning (CL) \cite{jaiswal2021CLsurvey1} has shown great potential in learning discriminative representations without labels.
The core idea of CL is to use the InfoNCE loss \cite{oord2018representationCPC_INFONCE} to measure how well a model can classify a feature representation from a set of unrelated negative samples.
For example, \cite{wu2018unsupervisedMemoryBank} learns feature representations at the instance level with a memory bank, where they try to maximally scatter instance embeddings over a unit sphere. 
MoCo \cite{he2020momentumMOCOv1} matches an encoded feature to a dynamic dictionary which evolves with a momentum updating strategy. 
SimCLR \cite{chen2020simpleSIMCLR} presents a simple framework that engenders negative samples from large mini-batches.

Further, Khosla \textit{et al.} \cite{khosla2020supervisedCL} extend the self-supervised contrastive approach to the fully-supervised setting. 
A supervised contrastive loss is put forward to effectively leverage label information, enabling intra-class compactness and inter-class dispersion of the feature space.
Wang \textit{et al.} \cite{wang2021denseCL} propose dense contrastive learning that works at pixel-level and achieves superior performance than MoCo on downstream dense prediction tasks.
In addition, Li \textit{et al.} \cite{prototypeCL} propose to facilitate contrastive learning with clustering.
They propose ProtoNCE loss that absorbs the advantages of contrastive learning and cluster-based unsupervised representation method \cite{tian2017deepcluster}, showing substantial improvements on several benchmarks.  





\noindent \textbf{Contrastive Learning in Segmentation.}
Recently, many studies leverage contrastive learning to promote image segmentation.
These works utilize pixel-level or patch-level contrastive learning to improve semantic segmentation under fully supervised \cite{wang2021exploring_fully_supervised}, semi-supervised \cite{alonso2021semi1}, weakly-supervised \cite{SPML}, and unsupervised \cite{van2021unsupervised1} settings.
Caron \textit{et al.} \cite{SPML} improve WSSS by pixel-to-segment contrast, where they assume the segments are known in advance. In their work, they use SEAM to generate CAMs.
Instead, we build on top of existing WSSS methods and directly produce better quality seeds for segmentation.
Recently, some studies apply contrastive learning to domain adaptation \cite{liu2021domain1} and few-shot \cite{liu2021fewshot1} semantic segmentation, also showing impressive results.

\noindent \textbf{Consistency Regularization.}
Consistency regularization is a hot topic in the field of semi-supervised semantic segmentation.
The idea is to enforce semantic or distribution consistency between various perturbations, such as image augmentation \cite{ke2020guidedKe} and network perturbation \cite{zhang2020wcp, chen2021semiCPC}.
For example, Ke \textit{et al.} \cite{ke2020guidedKe} enforce cross probability consistency; Chen \textit{et al.} \cite{chen2021semiCPC} impose consistency regularization on two networks perturbed with different parameters for the same input image.
The common goal of such methods is to construct appropriate supervisions by imposing consistency regularization, which greatly improves segmentation performance of semi-supervised segmentation. 
However, this idea is rarely studied in the case of weakly supervised segmentation.
Our proposed cross-view pixel-to-prototype contrast can be regarded as imposing feature semantic consistency regularization across different views of each image.





\section{Methodology}
Our method can be interpreted as a regularization term that is adaptable to any existing WSSS framework without changing the inference procedure.
The overall loss function for training such a model is the linear combination of the cross-view contrastive loss $\mathcal{L}^{\text{cross}}$ and the intra-view contrastive loss $\mathcal{L}^{\text{intra}}$:
\begin{equation}
    \mathcal{L}^{\text{contrast}} = \alpha \mathcal{L}^{\text{cross}} + \beta \mathcal{L}^{\text{intra}}
\end{equation}
where $\alpha$, $\beta$ are two positive constants.

In this section, we first review how to generate CAMs, then introduce our proposed pixel-to-prototype contrast and how to estimate the prototypes, and finally elaborate on applying contrastive learning across different views and within per single view of each image.
The framework of our method is illustrated in \cref{overall}. 
Following the common practice, we first generate pixel-wise pseudo masks with our proposed method, then use them to train a DeepLab \cite{chen2014semantic_Deeplabv1, chen2017deeplabV2} segmentation network.

\subsection{Preliminary}
We begin with a brief review on how CAMs are generated by visualization technique.
Given a CNN (\textit{e.g.} ResNet38 \cite{Resnet38}), we denote the last convolutional feature maps by $\boldsymbol{f} \in \mathbb{R}^{D \times HW}$, where $HW$ is the spatial size and $D$ is the channel dimension.
A Global Average Pooling (GAP) operation is applied to aggregate the feature maps. Next, a fully connected layer with parameters $\boldsymbol{w} \in \mathbb{R}^{C \times D} $ is applied to retrieve the class scores. Here, $C$ is the number of classes. Formally, the score for class $c$ is obtained as
\begin{equation}
    s_c = \frac{1}{HW} \sum_{j=1}^{D}\boldsymbol{w}_{c,j}\sum_{i}\boldsymbol{f}_{j,i}
\end{equation}

After that, the CAM $\boldsymbol{m}_{c}$ for class $c$ is given by:
\begin{equation}
  \boldsymbol{m}_{c} = \text{ReLU}\left(\sum_{j=1}^D\boldsymbol{w}_{c, j}\boldsymbol{f}_{j,:} \right)
\end{equation}

It is worth noticing that a theoretically equivalent and much convenient way to compute CAMs is directly selecting the feature maps of the last convolutional layer \cite{zhang2018adversarialCAMfirst}. Given the feature maps $\boldsymbol{f}$, we can add an additional convolutional layer which consists of $C$ kernels with a size of $1 \times 1$ and stride $1$ on top of $\boldsymbol{f}$ to obtain $\boldsymbol{f}' \in \mathbb{R}^{C\times HW}$. The score for class $c$ is then computed by applying GAP to $\boldsymbol{f}'$. The maps $\boldsymbol{f}'$ followed by ReLU function is directly used as the CAMs. 
In this paper, we follow this way for computing the CAMs.

\subsection{Pixel-to-Prototype Contrast}
\begin{figure}[t]
\centering
\begin{center}
\includegraphics[width=0.48\textwidth]{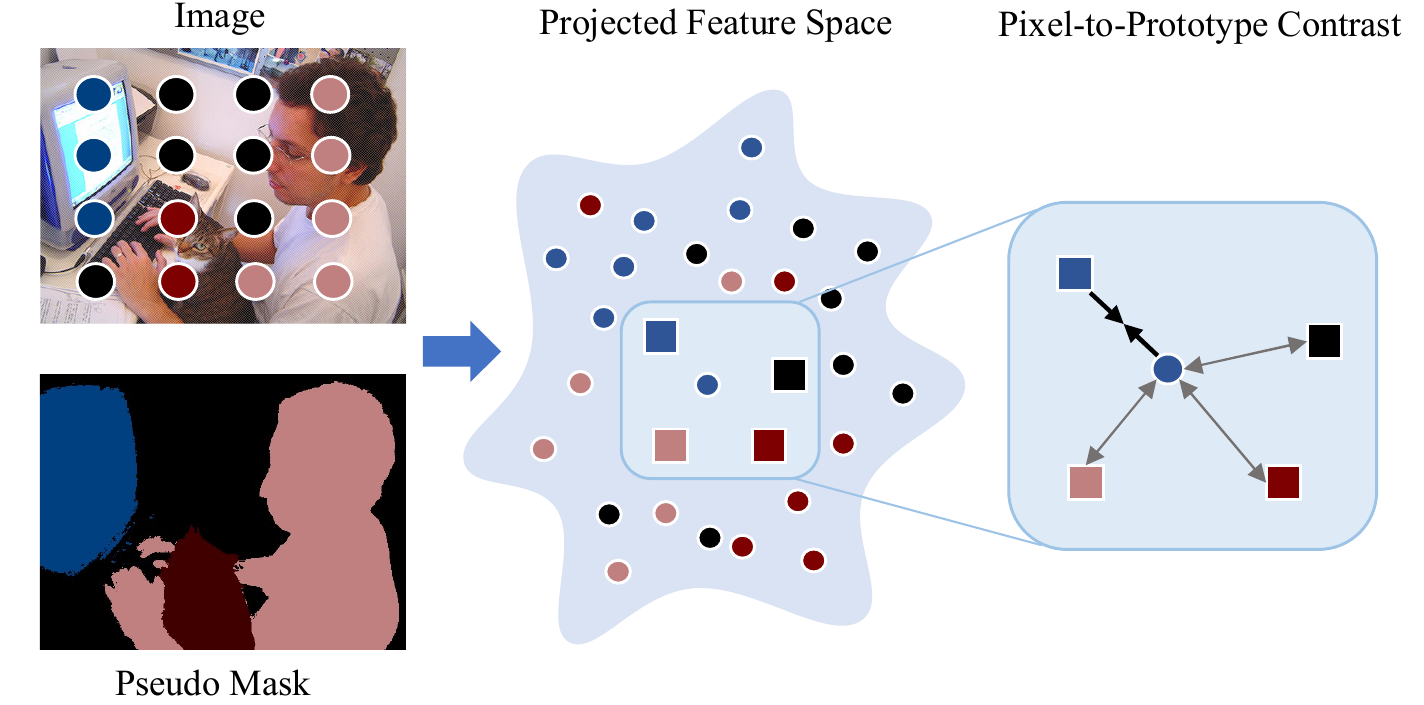}
\end{center}
\caption{Illustration of the pixel-to-prototype contrast in the projected feature space. Pixel embeddings (circles) and prototypes (squares) of the same color are of the same category.}
\label{fig:contrast}
\end{figure}

Given the CAM of an image, we use a pixel-wise argmax function to generate the pseudo mask $\boldsymbol{y}$, \textit{i.e.}, $\boldsymbol{y} = \text{argmax}(\boldsymbol{m})$, which determines the category of each pixel.
For each category, there exists a representative embedding, \textit{i.e.}, the prototypes, denoted by $\mathcal{P} = \{\boldsymbol{p}_{c} \}_{c=1}^C$.
Our goal is to learn discriminative feature embedding for each pixel aided by contrastive learning in a projected feature space.
The idea is shown in \cref{fig:contrast}.
We first obtain pixel-wise projected feature $\boldsymbol{v}_i \in \mathbb{R}^{128}$ by a projector, which is implemented with a $1\times 1$ convolutional layer followed by ReLU. 
Then, given $\boldsymbol{v_i}$ and $\mathcal{P}$, the pixel-to-prototype contrast $\mathcal{F}(\cdot)$ holds the following formulation:
\begin{equation}
    \mathcal{F}(\boldsymbol{v}_i;\ \boldsymbol{y}_i;\ \mathcal{P}) = -\text{log} \frac{\text{exp}(\boldsymbol{v}_i \cdot \boldsymbol{p}_{\boldsymbol{y}_i} / \tau)}{\sum_{\boldsymbol{p}_c \in \mathcal{P}} \text{exp}(\boldsymbol{v}_i \cdot \boldsymbol{p}_c / \tau) }
    \label{CL-Loss}
\end{equation}
where $\boldsymbol{y}_i \in [1,2,...,C]$ is the pseudo label of pixel $i$, which determines the positive prototype $\boldsymbol{p}_{\boldsymbol{y}_i}$.
$\tau$ is the temperature parameter, which is set to $0.1$ following a common practice.

\subsection{Prototype Estimation}
We further describe how we generate the prototypes.
A possible solution is to mine the pixel-level semantic structure through clustering, as done in unsupervised semantic segmentation \cite{picie}.
However, in the case of weakly supervised setting, this approach cannot make full use of image tag information and usually requires over-clustering to achieve better performance \cite{chang2020weaklySubCategory, tian2017deepcluster}. The resulting clusters generally cannot well match the true categories.

In this work, we treat the pixel-wise CAM values as the confidences and propose to estimate the prototypes from pixel-wise feature embeddings that are with the highest confidences.
Specifically, for all pixels assigned to class $c$,
we empirically choose the ones with top $K$ confidences to estimate the prototype.
The prototype $\boldsymbol{p}_c$ is computed as a weighted average of the projected pixel-wise embeddings:
\begin{equation}
    \boldsymbol{p}_c = \frac{\sum_{i \in \Omega_c} \boldsymbol{m}_{c,i} \boldsymbol{v}_{i}}{\sum_{i'\in \Omega_c} \boldsymbol{m}_{c,i'}}
\end{equation}
where $\Omega_c$ is the collection of top $K$ pixels of class $c$, and each pixel $i$ has CAM value $\boldsymbol{m}_{c, i}$.
A subsequent L2 normalization is applied to each prototype. Here, $K$ is a hyper-parameter and a smaller $K$ means higher confidence for computing the prototype.

Moreover, in order to capture the global context of the entire dataset, we compute prototypes across the training batch, that is, choosing pixels with the highest CAM values in the whole training batch.


\subsection{Cross-view Contrast}
Given the formulation of pixel-to-prototype contrastive loss shown in \cref{CL-Loss}, we describe in detail how we apply cross-view contrast under the guidance of the cross-view semantic consistency.
Specifically, given an image as the \textit{source} view $S$, we generate a \textit{target} view $T$ by a spatial transformation $A(\cdot)$, as shown in \cref{overall}.
Then, the two views are encoded using a pre-trained CNN backbone, which are further processed to obtain two CAMs.
We apply the same transformation $A(\cdot)$ to the feature map and the CAM of the source view, with the same purpose as \cite{wang2020selfSEAM}.

\textbf{Cross Prototype Contrast.}
Considering that there should be semantic consistency between the two views, the prototype from one view can serve as supervisory signal to another view, and vice versa.
Precisely, given a pixel $i$ with its pseudo label $\boldsymbol{y}_i \in [1,2,...,C]$ and the projected feature embedding $\boldsymbol{v}_i$,
the prototypes $\mathcal{P}'=\{\boldsymbol{p}'_{c}\}_{c=1}^C$ from another view are borrowed to impose regularization to the current view.
With the definition of pixel-to-prototype contrast shown in \cref{CL-Loss}, the cross prototype contrastive loss is calculated by
\begin{equation}
    \mathcal{L}^{\text{cp}} = \frac{1}{\lvert I \rvert}\sum_{i\in I}\mathcal{F}(\boldsymbol{v}_i;\ \boldsymbol{y}_i;\ \mathcal{P'})
\end{equation}
where $I$ denotes the whole image and $\lvert \cdot \rvert$ represents the cardinality.

\textbf{Cross CAM Contrast.}
Furthermore, the CAM from one view can be utilized to impose consistency regularization to another view as well.
The CAM determines the pseudo mask of a view.
Thus, for a pixel $i$ with prototypes $\mathcal{P}$ from its own view, we utilize the pseudo label $\boldsymbol{y}_i'$ from another view to determine the positive prototype and negative prototypes.
Analogously, the cross CAM contrastive loss can be written as:
\begin{equation}
    \mathcal{L}^{\text{cc}} = \frac{1}{|I|} \sum_{i \in I}\mathcal{F}(\boldsymbol{v}_i;\ \boldsymbol{y}'_i;\ \mathcal{P})
\end{equation}

It is noticeable that the cross-view contrast is symmetrical, since both \textit{source} view and \textit{target} view can serve as the current view for computing $\mathcal{L}^{\text{cp}}$ and $\mathcal{L}^{\text{cc}}$.
Eventually, the respective $\mathcal{L}^{\text{cp}}$ and $\mathcal{L}^{\text{cc}}$ of the two views are added together as the total cross-view contrastive loss $\mathcal{L}^{\text{cross}}$.
For simplicity, we only present the formulation of one view as follows:
\begin{equation}
    \mathcal{L}^{\text{cross}} = \mathcal{L}^{\text{cp}} + \mathcal{L}^{\text{cc}}
\end{equation}

\subsection{Intra-view Contrast}
\textbf{Intra-view Contrast.} 
According the second hypothesis of intra-class compactness and inter-class dispersion, we further propose intra-view contrast that is conducted within per single view of each image.
As opposed to the cross-view contrast, for a pixel $i$ with pseudo label $\boldsymbol{y}_i$, the intra-view contrast takes prototypes $\mathcal{P}$ from the current view to perform pixel-to-prototype contrastive learning:
\begin{align}
    \mathcal{L}^{\text{intra}} &= \frac{1}{|I|} \sum_{i\in I}  \mathcal{F}(\boldsymbol{v}_{i};\  \boldsymbol{y}_{i};\ \mathcal{P})
    \label{aaa}
\end{align}

The intra-view contrast is conducted on both views, and we do not list the symmetrical form in \cref{aaa} for simplicity.
However, we experimentally find that trivially introducing $\mathcal{L}^{\text{intra}}$ could cause performance degeneration. 
The reason is that there are no precise pixel-wise annotations in the case of weakly supervised setting, the pseudo label $\boldsymbol{y}_{i}$ assigned to pixel $i$ could be inaccurate, resulting in inaccurate contrasts.
Motivated by the hard example mining strategies in contrastive learning \cite{CLwith_hard_negatives, wang2021exploring_fully_supervised}, we alleviate this issue by introducing semi-hard prototype mining.
Moreover, we also adopt a hard pixel sampling strategy to focus more on pixel samples hard for segmentation.

\textbf{Semi-hard Prototype Mining.} For a pixel $i$, the assigned label $\boldsymbol{y}_{i}$ determines the positive prototype $\boldsymbol{p}_{\boldsymbol{y}_{i}}$ and negative prototypes $\mathcal{P}_N = \mathcal{P}\backslash \boldsymbol{p}_{\boldsymbol{y}_{i}}$.
Inspired by \cite{wang2021exploring_fully_supervised}, rather than directly using $\mathcal{P}_N$, we adopt semi-hard prototype mining: for each pixel, we first collect the top $60\%$ hardest negative prototypes, from which we choose $50\%$ as the negative samples to compute the intra-view contrastive loss.

Here, a remaining question is how to define `harder' prototypes. Following \cite{wang2021exploring_fully_supervised}, for pixel $i$, we view the prototypes except $\boldsymbol{p}_{\boldsymbol{y}_{i}}$ with dot products to pixel feature embedding $\boldsymbol{v}_{i}$ closer to $1$ to be harder, \textit{i.e.}, prototypes that are similar to the pixel. 


\textbf{Hard Pixel Sampling.} We also introduce hard pixel sampling to make better use of hard pixels.
Specially, instead of using all pixels belonging to a prototype $\boldsymbol{p}_{c}$ to calculate the intra-view contrastive loss, we adopt a per-class pixel sampling strategy: for each class, half of the pixels are randomly sampled and half are the hard ones.

Unlike \cite{wang2021exploring_fully_supervised}, in this part, we define `harder' pixels without ground truth during training.
For a prototype $\boldsymbol{p}_c$, we view the belonging pixels with dot products to $\boldsymbol{p}_c$ closer to $-1$ to be harder, \textit{i.e.}, pixels that are dissimilar to the prototype.
The definition of `harder' pixels is exactly the opposite of `harder' prototypes, as the pixel far away from the corresponding prototype requires more attention to be pulled closer to the prototype in order to improve intra-class compactness.

We experimentally prove that equipped with the two strategies, we mitigate the effect of incorrect contrasts and make better use of hard examples, which further improves performance.

\section{Experiment}
\subsection{Datasets and Baselines}
\textbf{Datasets.} We evaluate our proposed method on PASCAL VOC 2012 segmentation dataset \cite{everingham2010pascalVOC2012}, the standard benchmark for WSSS.
The dataset consists of 21 classes including a background, with 1,464, 1,449, and 1,456 images for train, validation, and test set, respectively. Following the common practice in semantic segmentation, we use the augmented train set that consists of 10,582 images \cite{hariharan2011semanticVOCtrainaug} for training.
We report the mean Intersection-over-Union (mIoU) for evaluation, and the mIoU on the VOC test set is obtained from the official evaluation server. 


\textbf{Baselines.} We choose two strong models, SEAM \cite{wang2020selfSEAM} and EPS \cite{lee2021railroadEPS} as our baselines. SEAM proposes a CAM equivariant regularization to narrow the supervision gap.
EPS utilizes saliency maps as an additional supervision. They achieve state-of-the-art performance for WSSS. 
We build on top of these models to evaluate the effectiveness of our proposed method.

\subsection{Implementation Details}
Following SEAM and EPS, ResNet38 is adopted as the backbone network with output stride equals 8. The images are randomly rescaled to the range of $[448, 768]$ by the longest edge and then cropped by $448 \times 448$ as the input size of the network following \cite{wang2020selfSEAM}.
We use rescale transformation that resizes the source image to a size of $128 \times 128$, keeping the multiple of the output stride. This is slightly different from SEAM, but the rescaling degree is nearly the same. The CNN backbone and projector share weights between the two views. The projected features used for contrastive loss have a dimension of $128$. When imposing our proposed contrastive regularization $\mathcal{L}^{contrast}$ to SEAM and EPS, we set $\alpha = 0.1$ and $\beta=0.1$ in order to keep balance with classification loss.
We follow the training and inference procedure in SEAM and EPS, including the training epoch, learning rate, learning rate decay policy, weight decay rate, and optimizer.



\begin{table}[tbp]
\small
\caption{Evaluation (mIoU (\%)) of the initial seed (Seed), the seed with CRF (+CRF), and the pseudo mask (Mask) refined by PSA \cite{ahn2018learningAffinityNet} on PASCAL VOC 2012 train set.}
\label{Tabel1: Affinity}
\centering
\rowcolors{11}{gray!20}{gray!10}
\begin{tabular}{l|c|c|c|c}
\hline
\textbf{Method}          & \textbf{Venue}   & \textbf{Seed} & \textbf{+CRF} & \textbf{Mask} \\
\hline \hline
PSA \cite{ahn2018learningAffinityNet}             & \multicolumn{1}{r|}{CVPR'18} & 48.0 & -    & 61.0 \\
Chang et al. \cite{chang2020weaklySubCategory}     & \multicolumn{1}{r|}{CVPR'20} & 50.9 & 55.3 & 63.4 \\
CONTA \cite{zhang2020causalCONTA}           & \multicolumn{1}{r|}{NIPS'20} & 56.2 & 65.4 & 66.1 \\
EDAM \cite{wu2021embeddedEDMA}            & \multicolumn{1}{r|}{CVPR'21} & 52.8 & 58.2 & 68.1 \\
AdvCAM \cite{lee2021antiADVCAM}         & \multicolumn{1}{r|}{CVPR'21} & 55.6 & 62.1 & 68.0 \\
ECS-Net \cite{Sun_2021_ICCV_ECSNet}                              & \multicolumn{1}{r|}{ICCV'21} & 56.6 & 58.6 & - \\
OC-CSE \cite{kweon2021unlocking0C-CSE}                               & \multicolumn{1}{r|}{ICCV'21} & 56.0 & 62.8 & 66.9 \\
CPN \cite{zhang2021complementaryCPN}         & \multicolumn{1}{r|}{ICCV'21} & 57.4 & - & - \\
CDA \cite{Su_2021_ICCV_CDA}                                       & \multicolumn{1}{r|}{ICCV'21} & 58.4 & - & 66.4 \\
\hline
\multicolumn{5}{l}{\textit{\textbf{Improvement over baseline:}}}        \\
SEAM \cite{wang2020selfSEAM}        & \multicolumn{1}{r|}{CVPR'20} & 55.4 & 56.8 & 63.6 \\
Ours w/ SEAM & -       & 61.5\scriptsize{\color{red}{+6.1}} & 64.0\scriptsize{\color{red}{+7.2}} & 69.2\scriptsize{\color{red}{+5.6}} \\
EPS \cite{lee2021railroadEPS}         & \multicolumn{1}{r|}{CVPR'21} & 69.5 & 71.4 & 71.6 \\
Ours w/ EPS  & -       & \textbf{70.5}\scriptsize{\color{red}{+1.0}} &\textbf{73.3}\scriptsize{\color{red}{+1.9}} & \textbf{73.3}\scriptsize{\color{red}{+1.7}}   \\
\hline
\end{tabular}
\end{table}

\begin{table}[tbp]
\caption{Evaluation (mIoU (\%)) of the initial seed (Seed), the seed with CRF (+CRF), and the pseudo mask (Mask) refined by IRN \cite{ahn2019weaklyIRNet} on PASCAL VOC 2012 train set.}
\small
\label{Table2: IRNet}
\centering
\rowcolors{6}{gray!10}{gray!20}
\begin{tabular}{l|c|c|c|c}
\hline
\textbf{Method}          & \textbf{Venue}   & \textbf{Seed} & \textbf{+CRF} & \textbf{Mask} \\
\hline \hline
IRN \cite{ahn2019weaklyIRNet}            & \multicolumn{1}{r|}{CVPR'19} & 48.8 & 54.3 & 66.3 \\
MBMNet \cite{liu2020weaklyMBMNet}         & \multicolumn{1}{r|}{MM'20}   & 50.2 & -    & 66.8 \\
CONTA \cite{zhang2020causalCONTA}          & \multicolumn{1}{r|}{NIPS'20} & 48.8 & -    & 67.9 \\
AdvCAM \cite{lee2021antiADVCAM}         & \multicolumn{1}{r|}{CVPR'21} & 55.6 & 62.1 & 69.9 \\ 
\hline
Ours w/ SEAM & -       & \textbf{61.5} & \textbf{64.0} &\textbf{70.1} \\
\hline
\end{tabular}
\end{table}

After generating the pseudo masks, we train three semantic segmentation networks to make fair comparisons with the baseline models. Concretely, DeepLab-LargeFOV with ResNet38 is trained to make comparison with SEAM; DeepLab-LargeFOV and DeepLab-ASPP with ResNet101 are trained to make comparison with EPS, respectively.
During inference, we adopt multi-scale and flip operations as done in previous works.
Standard dense CRF is used as a post-processing procedure to refine the final segmentation masks. 

\subsection{Seed and Pseudo Mask Evaluation}
\begin{figure}[t]
\centering
\begin{center}
\includegraphics[width=0.48\textwidth]{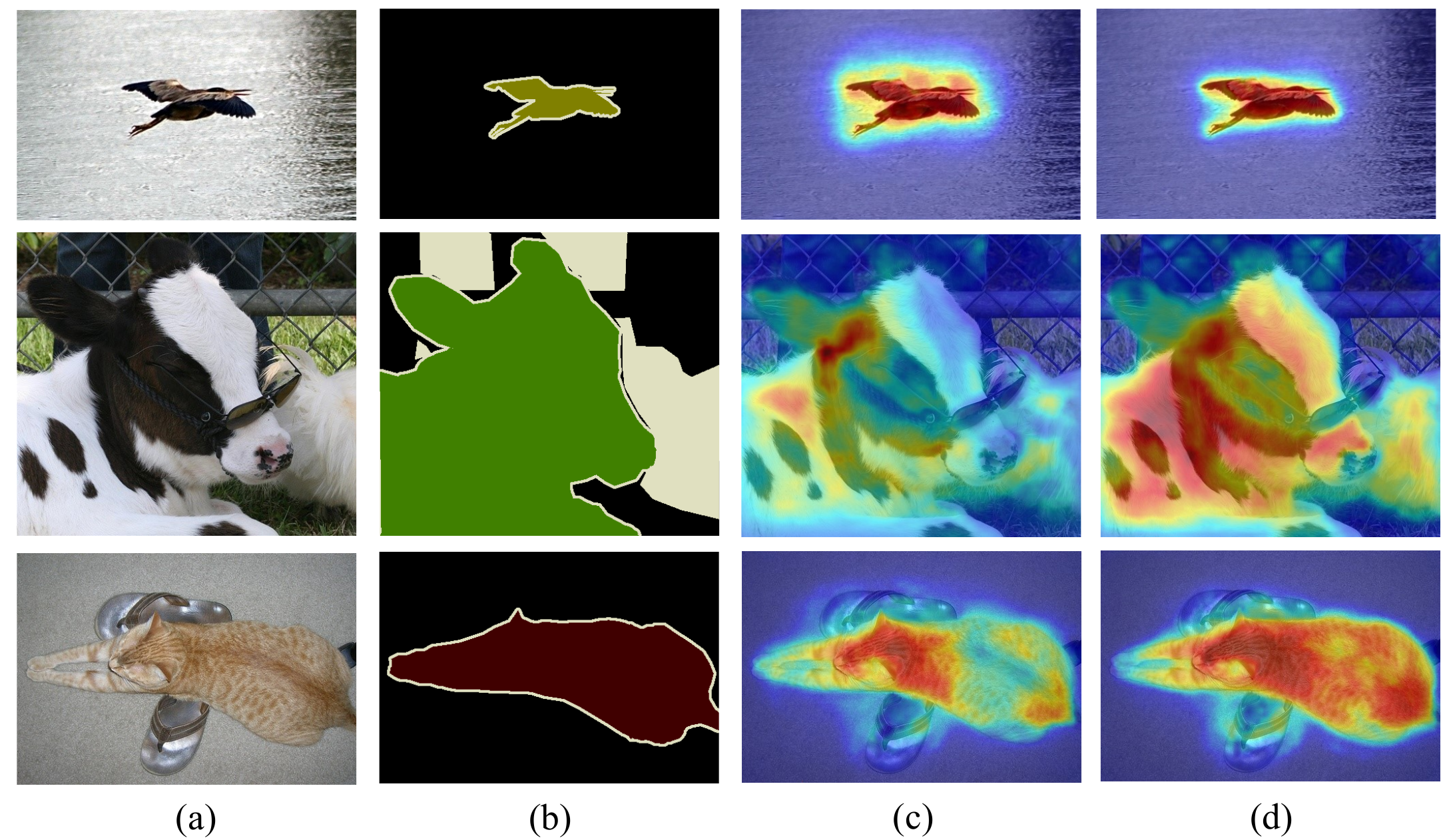}
\end{center}
\caption{The visualization of CAMs. (a) Images. (b) GT masks. (c) CAMs produced by SEAM. (d) CAMs produced by our method. Our method generates better CAMs than SEAM in terms of both accuracy and completeness.}
\label{fig:CAM}
\end{figure}

To verify the effectiveness of our method, we report the quantitative qualities of both initial seeds and pseudo masks on VOC. Following SEAM, the seeds are obtained by directly applying a range of thresholds to separate the foregrounds and backgrounds in the CAMs. The results are shown in \cref{Tabel1: Affinity}.
As can be seen, we improve SEAM by $6.0\%$ and $7.2\%$ mIoU on initial seed and seed+CRF, respectively, showing excellent performance.
\Cref{fig:CAM} indicates that the CAMs generated by our method not only cover the target objects completely but also show accurate boundaries.
Our generated CAMs are more accurate to match the ground truth segmentation masks than the baseline.
Moreover, compared with the recent methods CONTA \cite{zhang2020causalCONTA}, CPN \cite{zhang2021complementaryCPN}, and CDA \cite{Su_2021_ICCV_CDA} that built on top of SEAM, our method outperforms them by large margins.
A considerable improvement is also observed when applying our method to EPS.
Precisely, our method with EPS achieves $73.3\%$ mIoU on seed+CRF, achieving state-of-the-art performance.

A typical pipeline in WSSS is refining the initial seed by region growing with a random walk strategy.
Most methods refine their initial seeds with PSA \cite{ahn2018learningAffinityNet} or IRN \cite{ahn2019weaklyIRNet}.
Therefore, we also compare the qualities of the refined pseudo masks obtained by our method and other recent techniques. 
The 5th column of \cref{Tabel1: Affinity} shows the mIoU performance on the pseudo mask refined by PSA.
It can be observed that our method surpasses existing methods by large margins. Note that the seed quality of EPS with our method is high, we do not further refine it with PSA. Instead, the seed+CRF is directly supplied as the pseudo mask.

\Cref{Table2: IRNet} compares the performance between our method (with SEAM) and other counterparts refined using IRN.
Our method performs remarkably better than the best performing counterpart AdvCAM \cite{lee2021antiADVCAM}, exceeding it by a large margin of $5.9\%$ mIoU on the initial seed. As shown in the table, the pseudo mask performance gain of AdvCAM comes mainly from the refining procedure with IRN, while ours relies more on the high quality initial seed generated by our method.
Nevertheless, our method achieves state-of-the-art performance under the IRN refinement settings.


\begin{table}[tp]
\centering
\caption{Segmentation performance (mIoU (\%)) on Pascal VOC val and test sets using DeepLab-LargeFOV. The best result is marked in bold and the improvements over baseline model is marked in red. \textbf{$S$} means method using saliency maps.}
\label{V1}
\rowcolors{13}{gray!20}{gray!10}
\small
\begin{tabular}{l|c|c|c|c}
\hline
\textbf{Method}  & \textbf{Backbone} & \textbf{$S$} & \textbf{val} & \textbf{test} \\
\hline \hline

SEC \cite{kolesnikov2016SEC} \tiny{ECCV'16}        & VGG16             &  \checkmark          & 50.7         & 51.1          \\
MDC \cite{wei2018revisitingMDC} \tiny{CVPR'18}        & VGG16             &  \checkmark          & 60.4         & 60.8          \\
MCOF \cite{wang2018weaklyiterrefine1} \tiny{CVPR'18}        & ResNet101         &  \checkmark          & 60.3         & 61.2          \\
SeeNet \cite{hou2018selfErasing} \tiny{NIPS'18}        & ResNet101         &  \checkmark          & 63.1         & 62.8          \\
Lee \textit{et al.} \cite{lee2019frametoframe}  \tiny{ICCV'19}        & ResNet101         &  \checkmark          & 66.5         & 67.4          \\
OAA+ \cite{jiang2019integralOAA}     \tiny{ICCV'19}        & ResNet101         &  \checkmark          & 65.2         & 66.4          \\
CIAN \cite{fan2020cianCIAN}     \tiny{AAAI'20}        & ResNet101         &  \checkmark          & 64.3         & 65.3          \\
MCIS \cite{sun2020miningcrossimage}      \tiny{ECCV'20}        & ResNet101         &  \checkmark          & 66.2         & 66.9          \\
ICD  \cite{fan2020learningICD}     \tiny{CVPR'20}        & ResNet101         &  \          & 67.8         & 68.0          \\
ECS-Net \cite{Sun_2021_ICCV_ECSNet}                        \tiny{ICCV'21}        & ResNet38    & \ & 66.6 & 67.6 \\
Xu \textit{et al.}\cite{xu2021leveragingAuxSegNet}   \tiny{ICCV'21}        & ResNet38          &  \checkmark          & 69.0         & 68.6          \\
\hline
SEAM \cite{wang2020selfSEAM}        \tiny{CVPR'20}        & ResNet38          &  \          & 64.5         & 65.7          \\
Ours w/ SEAM              & ResNet38          &  \          & 67.7\scriptsize{\color{red}{+3.2}}         &  67.4\scriptsize{\color{red}{+1.7}}             \\
EPS \cite{lee2021railroadEPS}      \tiny{CVPR'21}        & ResNet101         &  \checkmark          & 71.0         & 71.8          \\
Ours w/ EPS                  & ResNet101         &  \checkmark          & \textbf{72.3}\scriptsize{\color{red}{+1.3}}         & \textbf{73.5}\scriptsize{\color{red}{+1.7}}             \\
\hline
\end{tabular}
\end{table}

\begin{table}[tp]
\centering
\caption{Segmentation performance (mIoU (\%) on Pascal VOC val and test sets using DeepLab-ASPP. The best result is marked in bold and the improvements over baseline model is marked in red. \textbf{$S$} means method using saliency maps.}
\label{Table:V2}
\rowcolors{19}{gray!20}{gray!10}
\small
\begin{tabular}{l|c|c|c|c}
\hline
\textbf{Method}  & \textbf{Backbone} & \textbf{$S$} & \textbf{val} & \textbf{test} \\
\hline \hline
PSA \cite{ahn2018learningAffinityNet}   \tiny{CVPR'18}        & ResNet38          & \           & 61.7         & 63.2          \\
IRN \cite{ahn2019weaklyIRNet}         \tiny{CVPR'19}        & ResNet50          & \           & 63.5         & 64.8          \\
FlickrNet \cite{Lee_2019_CVPR_FlikleNet}      \tiny{CVPR'19}        & ResNet101         &  \checkmark          & 64.9         & 65.3          \\
Zhang \textit{et al.} \cite{zhang2020splitting}  \tiny{ECCV'20}        & ResNet50          &  \checkmark          & 66.6         & 66.7          \\
Fan \textit{et al.} \cite{fan2020employing}   \tiny{ECCV'20}        & ResNet101         &  \checkmark          & 67.2         & 66.7          \\
Chen \textit{et al.} \cite{chen2020weaklyboundary} \tiny{ECCV'20}        & ResNet101         & \          & 65.7         & 66.6          \\
Chang \textit{et al.} \cite{chang2020weaklySubCategory} \tiny{CVPR'20}        & ResNet101         & \          & 66.1         & 65.9          \\
CONTA \cite{zhang2020causalCONTA} \tiny{NIPS'20}        & ResNet101         & \          & 66.1         & 66.7          \\
SPML \cite{SPML}          \tiny{ICLR'21}        & ResNet101         & \          & 69.5         & 71.6          \\
AdvCAM \cite{lee2021antiADVCAM}  \tiny{CVPR'21}        & ResNet101         & \          & 68.1         & 68.0          \\
EDAM \cite{wu2021embeddedEDMA}        \tiny{CVPR'21}        & ResNet101         &  \checkmark          & 70.9         & 70.6          \\Yao \textit{et al.} \cite{yao2021nonsalient}   \tiny{CVPR'21}        & ResNet101         &  \checkmark          & 68.3         & 68.5          \\
DRS \cite{kim2021discriminativeDRS}  \tiny{AAAI'21}        & ResNet101         & \          & 71.2         & 71.4          \\
Li \textit{et al.} \cite{li2021group}     \tiny{AAAI'21}        & ResNet101         &  \checkmark          & 68.2         & 68.5          \\
WSGCN \cite{pan2021weaklyGCN_ICME}             \tiny{ICME'21}        & ResNet101         & \          & 68.7         & 69.3          \\
CDA \cite{Su_2021_ICCV_CDA} \tiny{ICCV'21}   & ResNet38 & \ & 66.1 & 66.8 \\
CPN \cite{zhang2021complementaryCPN}      \tiny{ICCV'21}        & ResNet38          & \          & 67.8         & 68.5          \\
\hline
EPS \cite{lee2021railroadEPS}      \tiny{CVPR'21}        & ResNet101         &  \checkmark          & 70.9         & 70.8          \\
Ours w/ EPS                   & ResNet101         &  \checkmark          & \textbf{72.6}\scriptsize{\color{red}{+1.7}}         & \textbf{73.6}\scriptsize{\color{red}{+2.8}}  \\
\hline
\end{tabular}
\end{table}

\subsection{Segmentation Performance}
Generally, the generated pseudo masks are used to train a semantic segmentation network in a fully supervised manner.
To make a fair comparison, we report DeepLab-LargeFOV and Deeplab-ASPP segmentation performance of our method and compare them with existing methods in \cref{V1} and \cref{Table:V2}, respectively.
The SEAM trains a DeepLab-LargeFOV network with ResNet38 as the backbone, achieving $64.5\%$ and $64.7\%$ mIoU on PASCAL VOC val and test sets.
With the same settings except equipped with our method, we increase the segmentation mIoU by $3.2\%$ and $1.7\%$ on val and test sets, showing substantial improvement.
Moreover, our method with SEAM even surpasses many models with more powerful backbones.
With DeepLab-LargeFOV and ResNet101, EPS equipped with our method achieves $72.3\%$ and $73.5\%$ mIoU on PASCAL VOC val and test sets. 

\begin{figure}[tp]
\centering
\begin{center}
\includegraphics[width=0.48\textwidth]{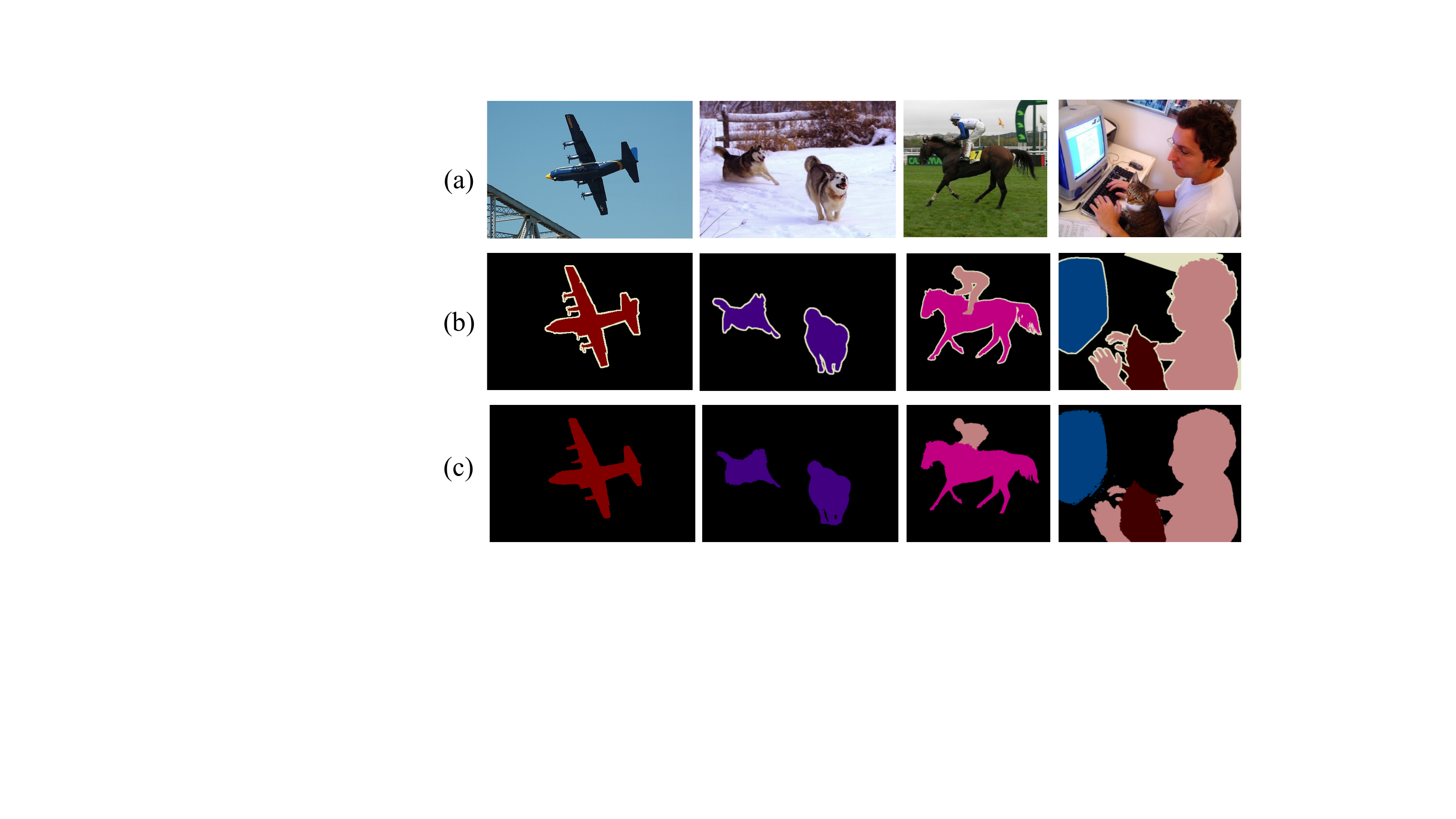}
\end{center}
\caption{Qualitative segmentation results on PASCAL VOC 2012 val set. (a) Images. (b) GT masks. (c) Segmentation masks predicted by DeepLab-ASPP (Ours w/ EPS).}
\label{fig:seg_results}
\end{figure}

In addition, we train a DeepLab-ASPP network with our generated pseudo masks. As shown in \cref{Table:V2}, our method outperforms all existing methods, achieving new state-of-the-art performance on PASCAL VOC 2012 benchmark. We present some segmentation results in \cref{fig:seg_results}, from which we can find that our method works well for images of both simple and challenging scenes. 

\begin{table}[tp]
\caption{Ablation performance (mIoU \%) of our proposed method. }
\label{Ablation1}
\centering
\small
\begin{tabular}{l|c|c|c|c}
\hline
\textbf{Method}      & \textbf{train} & \textbf{train}\scriptsize{+CRF} & \textbf{val} & \textbf{val}\scriptsize{+CRF} \\
\hline \hline
CAM                  & 47.43          & 52.40              & -            & -                \\
SEAM (baseline)    & 55.41          & 56.83              & 52.54        & 53.70            \\ 
\hline
\multicolumn{5}{l}{\textit{\textbf{Ablations:}}}                                             \\
$+$ Cross Prototype     & 59.08          & 61.98              & 55.75        & 58.33            \\
$+$ Cross CAM           & 60.35          & 63.10              & 57.33        & 59.68            \\
$+$ Intra View          & 59.80          & -                  & 56.97        & -                \\
$+$ Proto Mining        & 60.91          & 63.40              & 58.20        & 60.62            \\
$+$ Pixel Sampling      & 61.54          & 64.05              & 58.41        & 60.81            \\
\hline
\textbf{Improvement} &\textbf{+6.13 }          & \textbf{+7.22}               &\textbf{+5.87}         &\textbf{+7.11} \\
\hline
\end{tabular}
\end{table}

\subsection{Ablation Study}
To analyse how each component in our proposed method helps to improve WSSS, we present extensive ablation studies in this section. 
Here, all experiments are done with SEAM on PASCAL VOC 2012 dataset.

\textbf{Effectiveness of each component.} First, we demonstrate the effectiveness of each component. The results are shown in \cref{Ablation1}. As can be seen, with the two cross-view semantic consistency regularization term, \textit{i.e.} cross prototype contrast and cross CAM contrast, we improve the mIoU of SEAM from $55.41\%$ to $60.35\%$ on the train set, and from  $52.54\%$ to $57.35\%$ on the val set.
Furthermore, after applying the intra-view pixel-to-prototype contrast, we observe a slight performance drop on both train and val sets.
We adopt two sample mining strategies to mitigate this issue, by which we increase the mIoU by 1.74\% and 1.44\%.
Finally, with all these components, we increase the mIoU of SEAM by $6.13\%$ and $5.87\%$ on the train and val sets, showing significant improvements.

\textbf{Choices of $\boldsymbol{K}$.} We empirically choose $K$ pixel embeddings per-class in a mini-batch with the highest CAM values to evaluate the prototype.
A smaller $K$ indicates to use more confident pixels for estimation, while a larger $K$ could probably incur some wrong pixels from other categories. 
We conduct experiments to analyse how $K$ affects the initial seed performance.
As shown in \cref{Ablation2}, we test a wide range of $K$.
The results show that our proposed method is robust to the selection of $K$. Eventually, we set $K=32$ with the best performance.

\begin{table}[tbp]
\caption{Effects of $K$ for estimating the prototypes of our method. The results (mIoU (\%)) are reported on PASCAL VOC 2012 train set. $K$=$0$ means the SEAM baseline without contrast.}
\label{Ablation2}
\small
\centering
\begin{tabular}{c|c|c|c|c|c|c}
\hline
$K$   & 0 & 4 & 8 & 16 & 32 & 64 \\
\hline \hline
mIoU &  55.41  &  60.02 &  60.87  & 61.08   &\textbf{61.54} &  60.55   \\
\hline
\end{tabular}
\end{table}
\begin{table}[tp]
\caption{Ablations for more spatial transformations. The results (mIoU (\%)) are reported on PASCAL VOC 2012 train set.}
\small
\centering
\label{Ablation3}
\begin{tabular}{cccc|c}
\hline 
Rescale & Flip & Rotation & Translation & mIoU  \\
\hline \hline
        &      &          &             & 47.43 \\
\checkmark        &      &          &             &   61.54  \\
\checkmark         & \checkmark      &          &             & \textbf{61.63}  \\
\checkmark         &      & \checkmark          &             & 58.57  \\
\checkmark         &      &          & \checkmark             & 59.31  \\
\hline
\end{tabular}
\end{table}

\textbf{Spatial transformation.}
Similar to SEAM, the transformation in our method can be any spatial transformation.
In our implementation, we rescale the source image of size $448\times 448$ to $128 \times 128$ to get the target view in order to make sure the input size of the target view is a multiple of the output stride. 
This setting is slightly different from SEAM but remains almost identical to it.
Following SEAM, we also adopt three transformations to evaluate our method: random rotation in $[-20, 20]$ degrees, translation by $16$ pixels and horizontal flip.
The results are shown in \cref{Ablation3}. 
It can be seen that simply incorporating different transformations is not much effective, a similar observation to SEAM.

\section{Conclusion}
In this paper, we propose weakly-supervised pixel-to-prototype contrast that provides pixel-level supervisory signals to narrow the supervision gap and improve image-level WSSS. 
The pixel-to-prototype contrast is performed on both cross-view and intra-view of an image, which imposes cross-view feature semantic consistency regularization and facilitates intra(inter)-class compactness(dispersion) of the feature space.
Extensive experiments validate the superiority of our method.
In future, we will explore more prototype estimation ways and more application scenarios of our method, such as fully-supervised and semi-supervised segmentation.

\textbf{Acknowledgement.} The work was supported by the National Key Research and Development Program of China under Grant 2018YFB1701600, National Natural Science Foundation of China under Grant U20B2069.

{\small
\bibliographystyle{ieee_fullname}
\bibliography{egbib_simple}

\begin{thebibliography}{10}\itemsep=-1pt

\bibitem{ahn2019weaklyIRNet}
Jiwoon Ahn, Sunghyun Cho, and Suha Kwak.
\newblock Weakly supervised learning of instance segmentation with inter-pixel
  relations.
\newblock In {\em CVPR}, 2019.

\bibitem{ahn2018learningAffinityNet}
Jiwoon Ahn and Suha Kwak.
\newblock Learning pixel-level semantic affinity with image-level supervision
  for weakly supervised semantic segmentation.
\newblock In {\em CVPR}, 2018.

\bibitem{alonso2021semi1}
I\~nigo Alonso, Alberto Sabater, David Ferstl, Luis Montesano, and Ana~C.
  Murillo.
\newblock Semi-supervised semantic segmentation with pixel-level contrastive
  learning from a class-wise memory bank.
\newblock In {\em ICCV}, 2021.

\bibitem{bearman2016sPoints1}
Amy Bearman, Olga Russakovsky, Vittorio Ferrari, and Li Fei-Fei.
\newblock What’s the point: Semantic segmentation with point supervision.
\newblock In {\em ECCV}, 2016.

\bibitem{chang2020weaklySubCategory}
Yu-Ting Chang, Qiaosong Wang, Wei-Chih Hung, Robinson Piramuthu, Yi-Hsuan Tsai,
  and Ming-Hsuan Yang.
\newblock Weakly-supervised semantic segmentation via sub-category exploration.
\newblock In {\em CVPR}, 2020.

\bibitem{chen2014semantic_Deeplabv1}
Liang{-}Chieh Chen, George Papandreou, Iasonas Kokkinos, Kevin Murphy, and
  Alan~L. Yuille.
\newblock Semantic image segmentation with deep convolutional nets and fully
  connected crfs.
\newblock In {\em ICLR}, 2015.

\bibitem{chen2020weaklyboundary}
Liyi Chen, Weiwei Wu, Chenchen Fu, Xiao Han, and Yuntao Zhang.
\newblock Weakly supervised semantic segmentation with boundary exploration.
\newblock In {\em ECCV}, 2020.

\bibitem{chen2017deeplabV2}
Liang-Chieh Chen, George Papandreou, Iasonas Kokkinos, Kevin Murphy, and Alan~L
  Yuille.
\newblock Deeplab: Semantic image segmentation with deep convolutional nets,
  atrous convolution, and fully connected crfs.
\newblock {\em TPAMI}, 2017.

\bibitem{chen2020simpleSIMCLR}
Ting Chen, Simon Kornblith, Mohammad Norouzi, and Geoffrey Hinton.
\newblock A simple framework for contrastive learning of visual
  representations.
\newblock In {\em ICML}, 2020.

\bibitem{chen2021semiCPC}
Xiaokang Chen, Yuhui Yuan, Gang Zeng, and Jingdong Wang.
\newblock Semi-supervised semantic segmentation with cross pseudo supervision.
\newblock In {\em CVPR}, 2021.

\bibitem{picie}
Jang~Hyun Cho.
\newblock Picie: Unsupervised semantic segmentation using invariance and
  equivariance in clustering.
\newblock In {\em CVPR}, 2021.

\bibitem{everingham2010pascalVOC2012}
Mark Everingham, Luc Van~Gool, Christopher~KI Williams, John Winn, and Andrew
  Zisserman.
\newblock The pascal visual object classes (voc) challenge.
\newblock {\em IJCV}, 2010.

\bibitem{fan2020learningICD}
Junsong Fan, Zhaoxiang Zhang, Chunfeng Song, and Tieniu Tan.
\newblock Learning integral objects with intra-class discriminator for
  weakly-supervised semantic segmentation.
\newblock In {\em CVPR}, 2020.

\bibitem{fan2020employing}
Junsong Fan, Zhaoxiang Zhang, and Tieniu Tan.
\newblock Employing multi-estimations for weakly-supervised semantic
  segmentation.
\newblock In {\em ECCV}, 2020.

\bibitem{fan2020cianCIAN}
Junsong Fan, Zhaoxiang Zhang, Tieniu Tan, Chunfeng Song, and Jun Xiao.
\newblock Cian: Cross-image affinity net for weakly supervised semantic
  segmentation.
\newblock In {\em AAAI}, 2020.

\bibitem{hariharan2011semanticVOCtrainaug}
Bharath Hariharan, Pablo Arbel{\'a}ez, Lubomir Bourdev, Subhransu Maji, and
  Jitendra Malik.
\newblock Semantic contours from inverse detectors.
\newblock In {\em ICCV}, 2011.

\bibitem{he2020momentumMOCOv1}
Kaiming He, Haoqi Fan, Yuxin Wu, Saining Xie, and Ross Girshick.
\newblock Momentum contrast for unsupervised visual representation learning.
\newblock In {\em CVPR}, 2020.

\bibitem{hou2018selfErasing}
Qibin Hou, Peng{-}Tao Jiang, Yunchao Wei, and Ming{-}Ming Cheng.
\newblock Self-erasing network for integral object attention.
\newblock In {\em NIPS}, 2018.

\bibitem{huang2018weaklyRegionGrowing}
Zilong Huang, Xinggang Wang, Jiasi Wang, Wenyu Liu, and Jingdong Wang.
\newblock Weakly-supervised semantic segmentation network with deep seeded
  region growing.
\newblock In {\em CVPR}, 2018.

\bibitem{jaiswal2021CLsurvey1}
Ashish Jaiswal, Ashwin~Ramesh Babu, Mohammad~Zaki Zadeh, Debapriya Banerjee,
  and Fillia Makedon.
\newblock A survey on contrastive self-supervised learning.
\newblock {\em Technologies}, 2021.

\bibitem{jiang2019integralOAA}
Peng-Tao Jiang, Qibin Hou, Yang Cao, Ming-Ming Cheng, Yunchao Wei, and Hong-Kai
  Xiong.
\newblock Integral object mining via online attention accumulation.
\newblock In {\em ICCV}, 2019.

\bibitem{SPML}
Tsung{-}Wei Ke, Jyh{-}Jing Hwang, and Stella Yu.
\newblock Universal weakly supervised segmentation by pixel-to-segment
  contrastive learning.
\newblock In {\em ICLR}, 2021.

\bibitem{ke2020guidedKe}
Zhanghan Ke, Di Qiu, Kaican Li, Qiong Yan, and Rynson~WH Lau.
\newblock Guided collaborative training for pixel-wise semi-supervised
  learning.
\newblock In {\em ECCV}, 2020.

\bibitem{khosla2020supervisedCL}
Prannay Khosla, Piotr Teterwak, Chen Wang, Aaron Sarna, Yonglong Tian, Phillip
  Isola, Aaron Maschinot, Ce Liu, and Dilip Krishnan.
\newblock Supervised contrastive learning.
\newblock {\em NIPS}, 2020.

\bibitem{kim2021discriminativeDRS}
Beomyoung Kim, Sangeun Han, and Junmo Kim.
\newblock Discriminative region suppression for weakly-supervised semantic
  segmentation.
\newblock In {\em AAAI}, 2021.

\bibitem{kim2016deconvolutionalmulti-feature}
Hyo-Eun Kim and Sangheum Hwang.
\newblock Deconvolutional feature stacking for weakly-supervised semantic
  segmentation.
\newblock {\em arXiv:1602.04984}, 2016.

\bibitem{kolesnikov2016SEC}
Alexander Kolesnikov and Christoph~H Lampert.
\newblock Seed, expand and constrain: Three principles for weakly-supervised
  image segmentation.
\newblock In {\em ECCV}, 2016.

\bibitem{kweon2021unlocking0C-CSE}
Hyeokjun Kweon, Sung-Hoon Yoon, Hyeonseong Kim, Daehee Park, and Kuk-Jin Yoon.
\newblock Unlocking the potential of ordinary classifier: Class-specific
  adversarial erasing framework for weakly supervised semantic segmentation.
\newblock In {\em ICCV}, 2021.

\bibitem{Lee_2019_CVPR_FlikleNet}
Jungbeom Lee, Eunji Kim, Sungmin Lee, Jangho Lee, and Sungroh Yoon.
\newblock Ficklenet: Weakly and semi-supervised semantic image segmentation
  using stochastic inference.
\newblock In {\em CVPR}, 2019.

\bibitem{lee2019frametoframe}
Jungbeom Lee, Eunji Kim, Sungmin Lee, Jangho Lee, and Sungroh Yoon.
\newblock Frame-to-frame aggregation of active regions in web videos for weakly
  supervised semantic segmentation.
\newblock In {\em ICCV}, 2019.

\bibitem{lee2021antiADVCAM}
Jungbeom Lee, Eunji Kim, and Sungroh Yoon.
\newblock Anti-adversarially manipulated attributions for weakly and
  semi-supervised semantic segmentation.
\newblock In {\em CVPR}, 2021.

\bibitem{lee2021railroadEPS}
Seungho Lee, Minhyun Lee, Jongwuk Lee, and Hyunjung Shim.
\newblock Railroad is not a train: Saliency as pseudo-pixel supervision for
  weakly supervised semantic segmentation.
\newblock In {\em CVPR}, 2021.

\bibitem{prototypeCL}
Junnan Li, Pan Zhou, Caiming Xiong, and Steven C.~H. Hoi.
\newblock Prototypical contrastive learning of unsupervised representations.
\newblock In {\em ICLR}, 2021.

\bibitem{li2021group}
Xueyi Li, Tianfei Zhou, Jianwu Li, Yi Zhou, and Zhaoxiang Zhang.
\newblock Group-wise semantic mining for weakly supervised semantic
  segmentation.
\newblock In {\em AAAI}, 2021.

\bibitem{liu2021domain1}
Weizhe Liu, David Ferstl, Samuel Schulter, Lukas Zebedin, Pascal Fua, and
  Christian Leistner.
\newblock Domain adaptation for semantic segmentation via patch-wise
  contrastive learning.
\newblock {\em arXiv:2104.11056}, 2021.

\bibitem{liu2021fewshot1}
Weide Liu, Zhonghua Wu, Henghui Ding, Fayao Liu, Jie Lin, and Guosheng Lin.
\newblock Few-shot segmentation with global and local contrastive learning.
\newblock {\em arXiv:2108.05293}, 2021.

\bibitem{liu2020weaklyMBMNet}
Weide Liu, Chi Zhang, Guosheng Lin, Tzu-Yi Hung, and Chunyan Miao.
\newblock Weakly supervised segmentation with maximum bipartite graph matching.
\newblock In {\em MM}, 2020.

\bibitem{minaee2021imagesegmentation}
Shervin Minaee, Yuri~Y Boykov, Fatih Porikli, Antonio~J Plaza, Nasser
  Kehtarnavaz, and Demetri Terzopoulos.
\newblock Image segmentation using deep learning: A survey.
\newblock {\em TPAMI}, 2021.

\bibitem{oh2021backgroundBox2}
Youngmin Oh, Beomjun Kim, and Bumsub Ham.
\newblock Background-aware pooling and noise-aware loss for weakly-supervised
  semantic segmentation.
\newblock In {\em CVPR}, 2021.

\bibitem{oord2018representationCPC_INFONCE}
Aaron van~den Oord, Yazhe Li, and Oriol Vinyals.
\newblock Representation learning with contrastive predictive coding.
\newblock {\em arXiv:1807.03748}, 2018.

\bibitem{pan2021weaklyGCN_ICME}
Shun-Yi Pan, Cheng-You Lu, Shih-Po Lee, and Wen-Hsiao Peng.
\newblock Weakly-supervised image semantic segmentation using graph
  convolutional networks.
\newblock In {\em ICME}, 2021.

\bibitem{CLwith_hard_negatives}
Joshua~David Robinson, Ching{-}Yao Chuang, Suvrit Sra, and Stefanie Jegelka.
\newblock Contrastive learning with hard negative samples.
\newblock In {\em ICLR}, 2021.

\bibitem{shimoda2019self-supervised-difference}
Wataru Shimoda and Keiji Yanai.
\newblock Self-supervised difference detection for weakly-supervised semantic
  segmentation.
\newblock In {\em ICCV}, 2019.

\bibitem{Su_2021_ICCV_CDA}
Yukun Su, Ruizhou Sun, Guosheng Lin, and Qingyao Wu.
\newblock Context decoupling augmentation for weakly supervised semantic
  segmentation.
\newblock In {\em ICCV}, 2021.

\bibitem{sun2020miningcrossimage}
Guolei Sun, Wenguan Wang, Jifeng Dai, and Luc Van~Gool.
\newblock Mining cross-image semantics for weakly supervised semantic
  segmentation.
\newblock In {\em ECCV}, 2020.

\bibitem{Sun_2021_ICCV_ECSNet}
Kunyang Sun, Haoqing Shi, Zhengming Zhang, and Yongming Huang.
\newblock Ecs-net: Improving weakly supervised semantic segmentation by using
  connections between class activation maps.
\newblock In {\em ICCV}, 2021.

\bibitem{tang2018regularizedScrible2}
Meng Tang, Federico Perazzi, Abdelaziz Djelouah, Ismail Ben~Ayed, Christopher
  Schroers, and Yuri Boykov.
\newblock On regularized losses for weakly-supervised cnn segmentation.
\newblock In {\em ECCV}, 2018.

\bibitem{tian2017deepcluster}
Kai Tian, Shuigeng Zhou, and Jihong Guan.
\newblock Deepcluster: A general clustering framework based on deep learning.
\newblock In {\em Joint European Conference on Machine Learning and Knowledge
  Discovery in Databases}, 2017.

\bibitem{van2021unsupervised1}
Wouter Van~Gansbeke, Simon Vandenhende, Stamatios Georgoulis, and Luc Van~Gool.
\newblock Unsupervised semantic segmentation by contrasting object mask
  proposals.
\newblock In {\em ICCV}, 2021.

\bibitem{wang2021exploring_fully_supervised}
Wenguan Wang, Tianfei Zhou, Fisher Yu, Jifeng Dai, Ender Konukoglu, and Luc
  Van~Gool.
\newblock Exploring cross-image pixel contrast for semantic segmentation.
\newblock In {\em ICCV}, 2021.

\bibitem{wang2018weaklyiterrefine1}
Xiang Wang, Shaodi You, Xi Li, and Huimin Ma.
\newblock Weakly-supervised semantic segmentation by iteratively mining common
  object features.
\newblock In {\em CVPR}, 2018.

\bibitem{wang2021denseCL}
Xinlong Wang, Rufeng Zhang, Chunhua Shen, Tao Kong, and Lei Li.
\newblock Dense contrastive learning for self-supervised visual pre-training.
\newblock In {\em CVPR}, 2021.

\bibitem{wang2020selfSEAM}
Yude Wang, Jie Zhang, Meina Kan, Shiguang Shan, and Xilin Chen.
\newblock Self-supervised equivariant attention mechanism for weakly supervised
  semantic segmentation.
\newblock In {\em CVPR}, 2020.

\bibitem{wei2017objectERASING}
Yunchao Wei, Jiashi Feng, Xiaodan Liang, Ming-Ming Cheng, Yao Zhao, and
  Shuicheng Yan.
\newblock Object region mining with adversarial erasing: A simple
  classification to semantic segmentation approach.
\newblock In {\em CVPR}, 2017.

\bibitem{wei2018revisitingMDC}
Yunchao Wei, Huaxin Xiao, Honghui Shi, Zequn Jie, Jiashi Feng, and Thomas~S
  Huang.
\newblock Revisiting dilated convolution: A simple approach for weakly-and
  semi-supervised semantic segmentation.
\newblock In {\em CVPR}, 2018.

\bibitem{wu2021embeddedEDMA}
Tong Wu, Junshi Huang, Guangyu Gao, Xiaoming Wei, Xiaolin Wei, Xuan Luo, and
  Chi~Harold Liu.
\newblock Embedded discriminative attention mechanism for weakly supervised
  semantic segmentation.
\newblock In {\em CVPR}, 2021.

\bibitem{Resnet38}
Zifeng Wu, Chunhua Shen, and Anton van~den Hengel.
\newblock Wider or deeper: Revisiting the resnet model for visual recognition.
\newblock {\em Pattern Recognition}, 2019.

\bibitem{wu2018unsupervisedMemoryBank}
Zhirong Wu, Yuanjun Xiong, Stella~X Yu, and Dahua Lin.
\newblock Unsupervised feature learning via non-parametric instance
  discrimination.
\newblock In {\em CVPR}, 2018.

\bibitem{xu2021leveragingAuxSegNet}
Lian Xu, Wanli Ouyang, Mohammed Bennamoun, Farid Boussaid, Ferdous Sohel, and
  Dan Xu.
\newblock Leveraging auxiliary tasks with affinity learning for weakly
  supervised semantic segmentation.
\newblock In {\em ICCV}, 2021.

\bibitem{yao2020saliency2}
Qi Yao and Xiaojin Gong.
\newblock Saliency guided self-attention network for weakly and semi-supervised
  semantic segmentation.
\newblock {\em IEEE Access}, 2020.

\bibitem{yao2021nonsalient}
Yazhou Yao, Tao Chen, Guo-Sen Xie, Chuanyi Zhang, Fumin Shen, Qi Wu, Zhenmin
  Tang, and Jian Zhang.
\newblock Non-salient region object mining for weakly supervised semantic
  segmentation.
\newblock In {\em CVPR}, 2021.

\bibitem{zhang2020causalCONTA}
Dong Zhang, Hanwang Zhang, Jinhui Tang, Xian-Sheng Hua, and Qianru Sun.
\newblock Causal intervention for weakly-supervised semantic segmentation.
\newblock {\em NIPS}, 2020.

\bibitem{zhang2021complementaryCPN}
Fei Zhang, Chaochen Gu, Chenyue Zhang, and Yuchao Dai.
\newblock Complementary patch for weakly supervised semantic segmentation.
\newblock In {\em ICCV}, 2021.

\bibitem{zhang2020wcp}
Liheng Zhang and Guo-Jun Qi.
\newblock Wcp: Worst-case perturbations for semi-supervised deep learning.
\newblock In {\em CVPR}, 2020.

\bibitem{zhang2020splitting}
Tianyi Zhang, Guosheng Lin, Weide Liu, Jianfei Cai, and Alex Kot.
\newblock Splitting vs. merging: Mining object regions with discrepancy and
  intersection loss for weakly supervised semantic segmentation.
\newblock In {\em ECCV}, 2020.

\bibitem{zhang2018adversarialCAMfirst}
Xiaolin Zhang, Yunchao Wei, Jiashi Feng, Yi Yang, and Thomas~S Huang.
\newblock Adversarial complementary learning for weakly supervised object
  localization.
\newblock In {\em CVPR}, 2018.

\bibitem{zhou2016learningCAM}
Bolei Zhou, Aditya Khosla, Agata Lapedriza, Aude Oliva, and Antonio Torralba.
\newblock Learning deep features for discriminative localization.
\newblock In {\em CVPR}, 2016.

\end{thebibliography}
}

\end{document}